\documentclass[aps,pre,twocolumn,superscriptaddress,citesort]{revtex4-2}
\usepackage{graphicx}
\usepackage{amsmath}

\usepackage{bm}

\usepackage{hyperref}
\usepackage{xcolor}
\definecolor{tab_blue}{HTML}{1F77B4}
\hypersetup{%
  breaklinks=true,
  colorlinks=true,
  linkcolor=tab_blue,
  filecolor=tab_blue,
  urlcolor=tab_blue,
  citecolor=tab_blue,
}

\usepackage{lipsum}

\begin{document}

\title{DBSCAN in domains with periodic boundary conditions}
\author{Xander M. de Wit}
\email{x.m.d.wit@tue.nl}
\affiliation{Fluids and Flows group and J.M. Burgers Center for Fluid Mechanics, Department of Applied Physics and Science Education, Eindhoven University of Technology, 5600 MB Eindhoven, Netherlands}
\author{Alessandro Gabbana}
\affiliation{Fluids and Flows group and J.M. Burgers Center for Fluid Mechanics, Department of Applied Physics and Science Education, Eindhoven University of Technology, 5600 MB Eindhoven, Netherlands}

\date{January 22, 2025}

\begin{abstract}
Many scientific problems involve data that is embedded in a space with periodic boundary conditions. This can for instance be related to an inherent cyclic or rotational symmetry in the data or a spatially extended periodicity. When analyzing such data, well-tailored methods are needed to obtain efficient approaches that obey the periodic boundary conditions of the problem. In this work, we present a method for applying a clustering algorithm to data embedded in a periodic domain based on the DBSCAN algorithm, a widely used unsupervised machine learning method that identifies clusters in data. The proposed method internally leverages the conventional DBSCAN algorithm for domains with open boundaries, such that it remains compatible with all optimized implementations for neighborhood searches in open domains. In this way, it retains the same optimized runtime complexity of $\mathcal{O}(N\log N)$. We demonstrate the workings of the proposed method using synthetic data in one, two and three dimensions and also apply it to a real-world example involving the clustering of bubbles in a turbulent flow. The proposed approach is implemented in a ready-to-use Python package that we make publicly available.
\end{abstract}

\maketitle	

\section{Introduction}

Density-Based Spatial Clustering of Applications with Noise (DBSCAN) is a widely used unsupervised machine learning algorithm designed to identify clusters in spatial data by leveraging density-based criteria \cite{Ester1996,Schubert2017}. Unlike traditional clustering methods such as k-means, DBSCAN does not require prior knowledge of the number of clusters and is particularly effective at detecting clusters of arbitrary shapes and distinguishing noise points. The algorithm operates by grouping points that are closely packed together, based on a specified neighborhood radius $\epsilon$ and minimum number of points \verb|min_points| criteria. It is highly effective in applications ranging from geographical data analysis, to image segmentation, to many areas of physics \cite{Wibisono2021,Fauzi2021,Shen2016,Peng2018}.

Conventional implementations of the DBSCAN algorithm tacitly assume that all points reside in a space with open boundaries. There are, however, many applications where the embedding space of the data instead has periodic boundaries in some or all dimensions, as if the data points reside on the surface of a (possibly higher-dimensional) torus. Periodic boundary conditions, where particles exiting on one side of the domain re-enter on the other side of the domain, are a commonly used tool, particularly in computational physics simulations, to mimic systems that are spatially unbounded, as if extending infinitely in space. It is omnipresent, for example, in fluid dynamics or molecular dynamics simulations \cite{Allen1987}. However, it can also arise in many other areas when studying data that is naturally defined in modulo sense, such as angular data that periodically ranges from 0 to 360 degrees or the time of day in a 24-hour cycle.

Applying clustering algorithms in domains with periodic boundary conditions requires special care. While a well-tailored approach exists for clustering in periodic domains based on k-means clustering \cite{Miniak2022}, no such optimized implementation is publicly available for the DBSCAN clustering algorithm, to the best of the authors' knowledge. In this work, we discuss how to efficiently apply DBSCAN in domains with periodic boundary conditions.

Conceptually, one could achieve a DBSCAN with periodic boundary conditions simply by swapping out the conventional distance metric (e.g. Euclidean distance or Manhattan distance) for its periodic counterpart that takes into account the periodic boundary conditions when computing the distance between two points, as proposed for example in \cite{Turci2016}. In its most naive implementation, however, this would require $\mathcal{O}(N^2)$ operations to compute all pairwise distances. Instead, optimized implementations of nearest neighbor search algorithms achieve complexity of $\mathcal{O}(N\log N)$ or better by using some form of spatial indexing, such as the K-D tree or Ball tree algorithms \cite{Bentley1975,Friedman1977,Omohundro1989,Liu2006}. The approach we propose for clustering in domains with periodic boundaries remains fully compatible with existing optimized search algorithms designed for domains with open boundaries, ensuring efficient computation even for large datasets.


\section{Algorithm}
\begin{figure*}
    \centering
    \includegraphics[width=\linewidth]{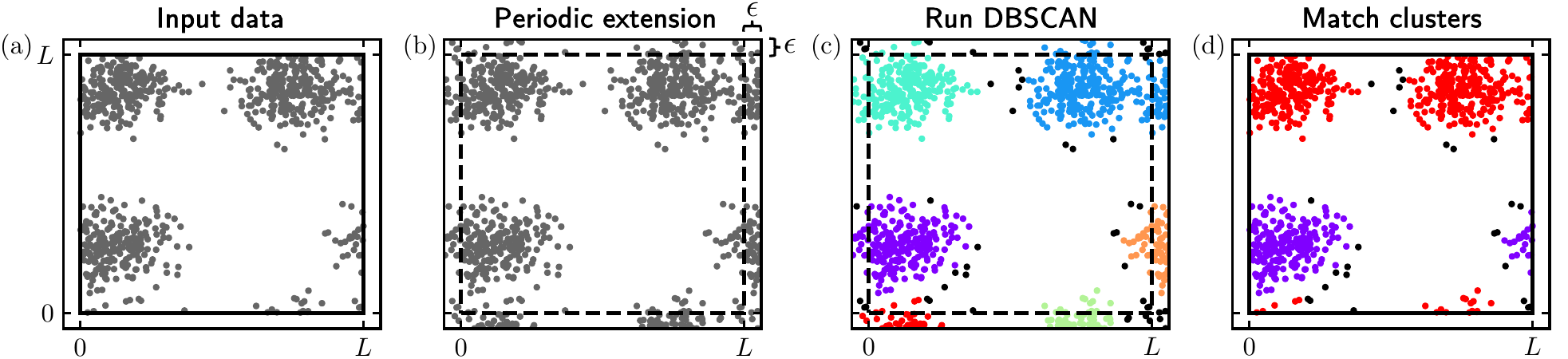}
    \caption{Example of the different steps of the algorithm for DBSCAN with periodic boundary conditions: (a) original input dataset, (b) periodic extension by $\epsilon$ (step 1), (c) DBSCAN of the extended dataset (step 2), (d) final clustering after linking and resolving equivalent clusters (steps 3 \& 4). This is a 2D example with periodicity $L$ and neighborhood $\epsilon = 0.06 L$.}
    \label{fig:demo_extd}
\end{figure*}
The approach we propose leverages the property of DBSCAN that proximity is defined by a single well-defined radius $\epsilon$. Consequently, the algorithm only needs to search for neighbors across the periodic boundary up to this distance.The method works by periodically extending the domain by a limited distance of $\epsilon$ in all periodic directions. This allows the clustering problem to be solved by applying the conventional DBSCAN algorithm -- designed for open boundaries -- to the extended domain. In the final step, the algorithm identifies and merges cluster labels assigned to different periodic copies of the same data point, ensuring that points in different periodic images are recognized as belonging to the same cluster in the periodic domain.

The algorithm takes as input the data points $\mathcal{S}$ (embedded in a space with dimension $D$) that need to be labeled, the lower and upper periodic boundaries \mbox{$\bm{x}_\textrm{min}=(x_\textrm{min}^{(1)}, x_\textrm{min}^{(2)}, ..., x_\textrm{min}^{(D)})$} and \mbox{$\bm{x}_\textrm{max}=(x_\textrm{max}^{(1)}, x_\textrm{max}^{(2)}, ..., x_\textrm{max}^{(D)})$}, respectively, and finally the DBSCAN parameters, being the neighborhood $\epsilon$ and \verb|min_points|. The procedure consists of four steps, which are illustrated with an example in Fig.~\ref{fig:demo_extd}:
\begin{enumerate}
    \item \emph{Periodic extension}. Extend data set $\mathcal{S}$ from \mbox{$[\bm{x}_\textrm{min}, \bm{x}_\textrm{max}]$} to \mbox{$[\bm{x}_\textrm{min}-\epsilon, \bm{x}_\textrm{max}+\epsilon]$} through periodic extension, saving the padded data points (the periodic copies) into $\mathcal{S}_\textrm{pad}$. For all padded points $s_\textrm{pad} \in \mathcal{S}_\textrm{pad}$, save the index of the corresponding point in the original dataset $\mathcal{S}$.
    \item \emph{DBSCAN}. Apply original DBSCAN with neighborhood $\epsilon$ and \verb|min_points| to all data points \mbox{$\mathcal{S}_\textrm{all} = \mathcal{S} \cup \mathcal{S}_\textrm{pad}$}, yielding labels $\mathcal{L}_\textrm{all}$.
    \item \emph{Linking equivalent clusters}. For each padded point $s_\textrm{pad} \in \mathcal{S}_\textrm{pad}$, compare its label $l_\textrm{pad}$ to the label of the corresponding point in the original dataset $l_\textrm{orig}$. If $l_\textrm{pad} \neq l_\textrm{orig}$, save the labels as a linked cluster if that link does not already exist. If one of the labels already exists in another link, extend that link by including the other label.
    \item \emph{Resolving linked clusters}. For all the saved linked clusters, replace the linked labels by a single unique label (e.g. the minimum of the linked labels). This yields the final labels $\mathcal{L}$ corresponding to the clustering of the original data points $\mathcal{S}$ obeying the periodic boundary conditions.
\end{enumerate}

Since this approach employs the conventional DBSCAN algorithm with open boundaries, it is automatically compatible with all optimized implementations of DBSCAN and its underlying neighbor search algorithms. Since the neighborhood distance $\epsilon$ is typically small with respect to the domain size, the number of padded points is typically a small fraction of the total number of points $N$. The impact on the performance of our approach for solving the clustering problem in the periodic domain is thus small with respect to the conventional clustering problem with open boundaries. And crucially, owing to its compatibility, it can be run at the same complexity of $\mathcal{O}(N\log N)$ that the optimized neighbor search algorithms for open boundaries are able to achieve.

\section{Implementation}
We have implemented the proposed approach for DBSCAN in domains with periodic boundaries in a Python package that is publicly available in the repository at \href{https://github.com/XanderDW/PBC-DBSCAN}{\texttt{github.com/XanderDW/PBC-DBSCAN}}. It uses the widely employed and highly optimized Scikit-learn implementation of DBSCAN \cite{scikit-learn} to ensure broad compatibility. The repository also provides ready-to-use code examples for the different example cases provided in this work.

\section{Examples with synthetic data}
\begin{figure}[b]
    \centering
    \includegraphics[width=\linewidth]{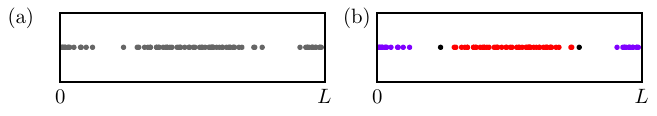}
    \caption{1D example of DBSCAN clustering with periodic boundary conditions with periodicity $L$ and neighborhood \mbox{$\epsilon=0.05 L$}. The example shows the raw data (a) and the clustering (b), where different colors represent different clusters, while black points indicate noise points that do not belong to a cluster.}
    \label{fig:demo-1D}
\end{figure}

Here we provide examples of the proposed approach for the DBSCAN clustering problem with periodic boundaries using data that is synthetically generated from (multivariate) Gaussian distributions.

Fig.~\ref{fig:demo-1D} depicts the simplest example of periodic clustering in one dimension. It shows that the algorithm successfully connects the purple cluster that traverses the periodic boundary.

\begin{figure}
    \centering
    \includegraphics[width=\linewidth]{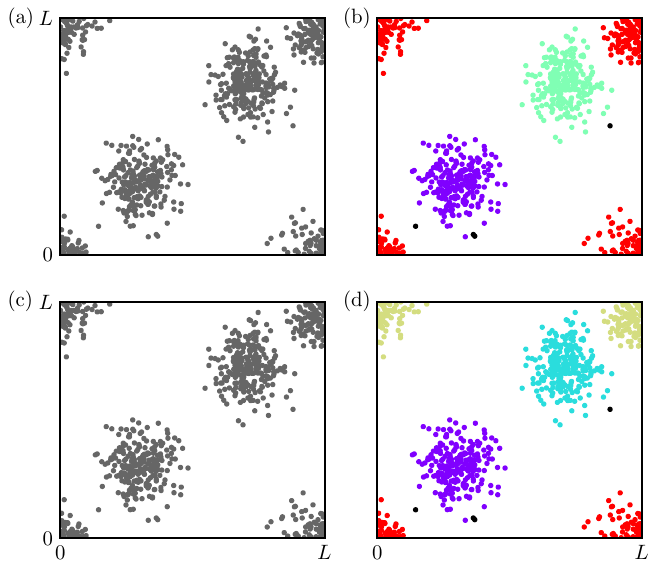}
    \caption{2D example of DBSCAN clustering with doubly periodic boundary conditions (a,b) and with singly periodic boundary conditions (c,d) where in the latter the left and right boundaries are periodic while the top and bottom boundaries are open. The periodicity is $L$ and neighborhood is $\epsilon = 0.08 L$. Panels and colors are as in Fig~\ref{fig:demo-1D}.}
    \label{fig:demo-2D}
\end{figure}

In Fig.~\ref{fig:demo-2D} we show an example in two dimensions, distinguishing the cases of doubly periodic Fig.~\ref{fig:demo-2D}(a,b) and singly periodic boundary conditions Fig.~\ref{fig:demo-2D}(c,d).

\begin{figure}[b]
    \centering
    \includegraphics[width=\linewidth]{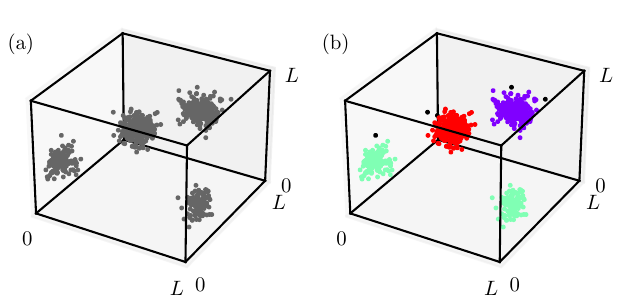}
    \caption{3D example of DBSCAN clustering with triply periodic boundary conditions with periodicity $L$ and neighborhood $\epsilon=0.08 L$. Panels and colors are as in Fig~\ref{fig:demo-1D}.}
    \label{fig:demo-3D}
\end{figure}

Finally, Fig.~\ref{fig:demo-3D} shows an example of periodic clustering in three dimensions, where all three dimensions have periodic boundaries.

Our implementation supports data with an arbitrary number of dimensions and can arbitrarily mix open boundaries and periodic boundaries for every dimension separately.

\section{Example with real data}
\begin{figure}[h]
    \centering
    \includegraphics[width=0.8\linewidth]{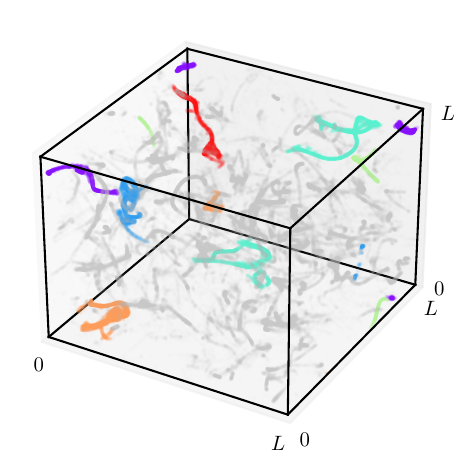}
    \caption{Example of DBSCAN clustering on a real dataset of light particles in turbulence in a 3D triply periodic domain with periodicity $L$ and neighborhood $\epsilon = 0.009 L$. Light particles tend to cluster in high-vorticity regions of the flow in filamentary structures. Colors shows the six largest clusters of particles as identified by the algorithm. Other clusters are colored in gray for readability.}
    \label{fig:demo-3D-real}
\end{figure}

Real world data can often involve clusters with highly non-Gaussian shapes. DBSCAN is very effective in identifying clusters of these complex shapes. One such example is encountered in turbulent flows, when studying the clustering of light bubbles submerged in a heavier turbulent fluid flow. There, bubbles are found to strongly concentrate in regions of high vorticity, forming filamentary clusters inside the cores of these elongated vortex structures \cite{Calzavarini2008,Toschi2009}. Such clustering behavior is typically studied computationally in domains with periodic boundary conditions to ensure full homogeneity and to eliminate any effect of confinement, such as boundary layer formation. An example is provided in Fig.~\ref{fig:demo-3D-real}, obtained from a direct numerical simulation of homogeneous isotropic turbulence with Lagrangian bubbles \cite{DeWit2024}. It shows that the clustering algorithm proposed in this work is able to successfully capture the bubble clusters in accordance with the periodic boundary conditions. Notice how, for instance, the turquoise cluster traverses the top/bottom boundary and the purple cluster crosses four different corners of the domain.

\section{Conclusions}
In this work, we have presented a clustering algorithm based on DBSCAN for data embedded in a domain with periodic boundaries. The approach leverages the conventional DBSCAN algorithm designed for open boundaries, ensuring compatibility with existing optimized neighborhood search methods. As a result, it maintains the same runtime complexity of $\mathcal{O}(N\log N)$ as conventional optimized DBSCAN algorithms. Our Python implementation of this method is publicly available as a ready-to-use package in the repository at \href{https://github.com/XanderDW/PBC-DBSCAN}{\texttt{github.com/XanderDW/PBC-DBSCAN}}.

\section*{Acknowledgments}
\vspace{-1mm}
This publication is part of the project “Shaping turbulence with smart particles” with Project No. OCENW.GROOT.2019.031 of the research program Open Competitie ENW XL which is (partly) financed by the Dutch Research Council (NWO).

\bibliography{main}

\end{document}